# CONCEPTUAL DESIGN GENERATION USING LARGE LANGUAGE MODELS

**Kevin Ma**
Dept. of Mechanical Engineering
University of California, Berkeley
Berkeley, CA, USA
kevinma1515@berkeley.edu

**Daniele Grandi**
Autodesk Research
San Francisco, CA, USA
daniele.grandi@autodesk.com

**Christopher McComb**
Dept. of Mechanical Engineering
Carnegie Mellon University
Pittsburgh, PA, USA
ccm@cmu.edu

**Kosa Goucher-Lambert**[*]
Dept. of Mechanical Engineering
University of California, Berkeley
Berkeley, CA, USA
kosa@berkeley.edu

## ABSTRACT
*Concept generation is a creative step in the conceptual design phase, where designers often turn to brainstorming, mindmapping, or crowdsourcing design ideas to complement their own knowledge of the domain. Recent advances in natural language processing (NLP) and machine learning (ML) have led to the rise of Large Language Models (LLMs) capable of generating seemingly creative outputs from textual prompts. The success of these models has led to their integration and application across a variety of domains, including art, entertainment, and other creative work. In this paper, we leverage LLMs to generate solutions for a set of 12 design problems and compare them to a baseline of crowdsourced solutions. We evaluate the differences between generated and crowdsourced design solutions through multiple perspectives, including human expert evaluations and computational metrics. Expert evaluations indicate that the LLM-generated solutions have higher average feasibility and usefulness while the crowdsourced solutions have more novelty. We experiment with prompt engineering and find that leveraging few-shot learning can lead to the generation of solutions that are more similar to the crowdsourced solutions. These findings provide insight into the quality of design solutions generated with LLMs and begins to evaluate prompt engineering techniques that could be leveraged by practitioners to generate higher-quality design solutions synergistically with LLMs.*

---

[*]Address all correspondence to this author.

## 1 INTRODUCTION

Research in engineering design points to the benefits of generating a large and diverse set of initial concepts during the early stage of design [1, 2, 3]. To support these efforts, research has begun to investigate the use of both human-powered (e.g., crowdsourcing) and computational methods that can quickly generate supplemental design concepts to support human design teams. For example, repositories of existing design knowledge have been leveraged to generate and retrieve variants of conceptual designs [4]. Other search systems involve using semantic networks to retrieve potential design solutions via the form of knowledge graphs that connects words and phrases of natural language with corresponding edges [5, 6, 7]. Also, design solutions can be crowdsourced from online crowdworkers as potential sources of inspiration for designers [8, 9].

In other domains, recent advances in ML have led to the development of large generative models that are capable of outputting creative results [10, 11]. In the design domain, some recent works have shown that ML models, such as variational autoencoders (VAEs) and generative adversarial networks (GANs), have the ability to output creative visual design solutions [12, 13], while LLMs have been recently used to output creative textual design solutions [14, 15].

In particular, Zhu *et al.* leveraged Generative Pre-Trained Transformers 3 (GPT-3), a specific LLM model, to generate 500 design solutions [14]. In [14], Zhu *et al.*, showed that GPT-3 was capable of generating novel and useful design solutions, and they laid out guidelines for defining concept generation tasks using LLMs. However, common metrics for evaluating design so-lutions, such as novelty and usefulness, are very hard to computationally assess due to their subjective nature and dependence on the design problem context.

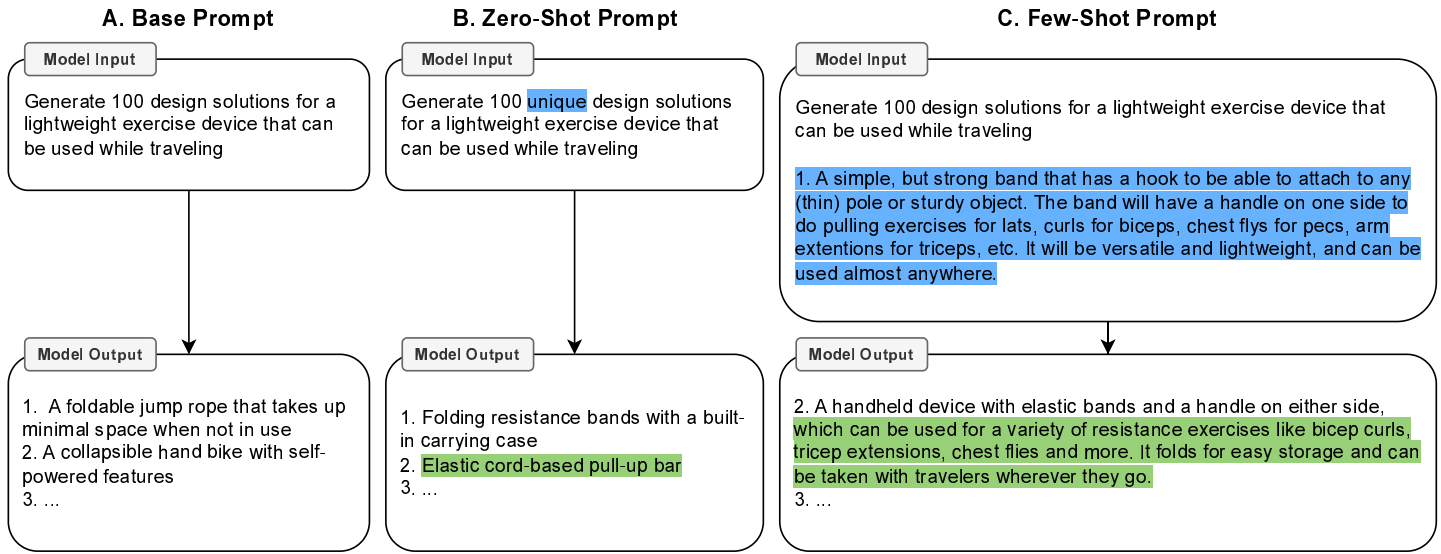

**FIGURE 1**: The text in *model input* is fed into the Large Language Model (LLM), which outputs 100 design solutions as previewed in *model output*. The **base prompt** (A) was also used to create the crowdsourced baseline dataset, while the **zero-shot prompt** (B) and the **few-shot prompt** (C) showcase the prompt engineering experiments performed to improve the generation of design solutions by the LLM.

In this work, we build on [14] by comparing design solutions generated with an LLM to a baseline of crowdsourced design solutions. We differentiate our methods from previous work by evaluating textual design solutions with computational metrics as well as human expert evaluations. Key differences between the crowdsourced solutions and LLM solutions, such as the diversity and uniqueness of solutions, are explored. The results are compared with expert evaluations of novelty, feasibility, and usefulness to assess the differences between computational and expert evaluations. Finally, we explore different methods of prompt engineering to quantitatively evaluate its impact on design solutions generated from LLMs (Figure 1). Using these research methods, this work specifically examines the following research questions (RQ):

1. In what ways do solutions generated with LLMs differ from those produced by crowdworkers?
2. How do computational metrics compare to human expert evaluations for assessing important characteristics of human and LLM-generated design solutions?
3. To what degree can prompt engineering be employed to vary the attributes of LLM-generated design solutions?

## 2 BACKGROUND

In this section we review works related to generating design solutions, pre-trained transformers, and different approaches for evaluating generated designs.

### 2.1 Generating Design Solutions

Design solutions are an effective means of inspirational stimuli for designers during the concept generation phase, but obtaining design solutions is challenging [16, 2, 17]. Crowdsourcing is a method to obtain design solutions [9] because it is an effective way of gathering creative ideas from a large pool of people, so designers can incorporate other creative examples into their own unique conceptual ideas [18]. Crowd workforces can be composed of untrained workers from online platforms such as Amazon Mechanical Turk and can generate large amounts of design solutions for trained designers, who can utilize the ideas as inspirational stimuli [8]. However, there exists a tradeoff between the quality and cost of solutions coming from either untrained workers or experienced designers [8].

Another method of obtaining design solutions is through visual design generations to support designers with the reuse of concepts [19]. Visual design solutions can be automatically generated using deep learning models such as Generative Adversarial Networks (GAN) and Variational Autoencoders (VAE) [20,21]. These generative models are capable of creating new designs and generalize novelties to a broad array of designs with no human interventions, which can aid designers during the synthesis and exploration phase [12]. Additionally, generative models are able to take in large corpus of existing designs to produce a wide range of new designs that are able to achieve functional goals of specific desired design problems [13]. Other forms of generative models make use of deep reinforcement learning algorithms to learn design strategies and construct design solutions by learning a policy that represents the agent's design strategy [22]. However, all these models need to be trained on large amounts of domain-specific data (which are expensive and scarce in the design domain) or are computationally expensive to train on a scale where they will be able to automate reasoning and understanding.

Recent developments in text-to-text autoregressive language models counteract this limitation by drawing in large amounts of data from a wide variety of sources. These LLMs are capable of inferring simple reasoning tasks based on the input provided and are capable of generating solutions given a simple task [23]. Research on LLM's application to design problems has been limited, however. The most recent work has shown that pre-trained LLMs, such as the generative pretrained transformers model series (GPT), are very capable of generating novel design concepts for the concept generation phase [14]. Experiments conducted by Zhu *et al.* reveal that GPT-3 models, the state-of-the-art GPT model, with few shot learning and fine-tuned GPT-2 models, an older GPT model, are able to receive input design problems and output design solutions with a high degree of novelty and usefulness [14, 15]. In this paper, we expand on prior work by evaluating design solutions generated with LLMs against a human baseline, using both computational and human expert metrics.

### 2.2 Pre-Trained Transformers

The field of natural langauge processing (NLP) explores how computers can use, understand, and manipulate natural language text or speech for various tasks, such as machine translation, question answering, and text summarization [24, 25, 26, 27]. Over the past decade, advances in deep learning models, including the development of transformers and self-attention mechanisms, have enabled the development of deep learning architectures that are highly effective and efficient at performing natural language generation [28, 29]. Recent studies have shown that scaling up both the size of the models and the corpus they are trained on achieves strong performances on many NLP tasks with minimal fine-tuning, leading to a proliferation of powerful LLMs capable of generating texts with minimal training time [11].

In this paper, we leverage a pre-trained generative model called GPT-3 to generate design solutions conditioned on inputs of design problems. GPT-3 is an autoregressive language model that utilizes the transformer architecture to produce human-like text [11, 28]. The model is trained on a large amount of data, containing billions of words scraped from the Internet and books. Recent development and advances in GPT-3 has shown that GPT-3 is capable of solving simple tasks at a similar level to human subjects and is able to outperform humans in decision-making tasks, such as multiarmed bandit tasks [23]. Additionally, GPT-3 has been broadly used as interaction mechanisms for coding and educational dialogue [30, 31]. Due to its wide applicability for a broad range of creative tasks and its ability to conduct simple reasoning and understanding, we utilize GPT-3 in this work to generate design solutions.

In addition to natural language generation, LLMs are capable of performing natural language understanding (NLU), which is primarily used to embed textual data into semantically meaningful vectors. Prior to the development of pre-trained transformers, embedding vectors were generated using non-contextual models that are capable of learning word and phrase representations (such as Word2Vec and Glove) [32, 33]. Recent developments in pre-trained transformers have led to the development of models that are able to contextualize words in a sentence and derive embedding values that are semantically meaningful conditioned on the context the words are being used in the sentence (such as ELMO and BERT) [34, 35]. Past works in the design communities have made use of non-contextual models to generate word and phrase representations for analysis [36, 5] but the use of state-of-the-art NLU models has been limited. In this paper, we experiment with SentenceBERT [37], a faster version of BERT, to embed textual design solutions to enable more accurate computational evaluations at a large scale.

### 2.3 Human and Computational Evaluation of Design Concepts

Creativity assessment methods to evaluate design concepts have been extensively studied in prior literature [38, 39, 40, 41, 42, 43, 44, 45]. Two of the most commonly used metrics are Consensual Assessment Technique (CAT) [46, 47, 38, 48] and the Shah, Vargas-Hernandez, and Smith (SVS) method [42, 49, 38, 50]. CAT, commonly used in design methodology literature, involves asking domain experts to subjectively rate designs on different criteria, such as creativity, novelty, or usefulness. On the other hand, the SVS method is comparatively less resource-demanding, but has been found to be less aligned with domain experts' opinions [51]. While the cost and subjectivity limitations that these methods present has been pointed out in various studies [52, 53, 54], they remain valuable techniques for evaluating design concepts [38]. In this paper, we leverage CAT to evaluate design concepts generated from both crowdsourced workers and GPT-3 models.

While human evaluation of design concepts remains the gold standard in the field [45], the introduction of ML methods for generating design concepts demands a more cost-effective and scalable evaluation method, more apt to evaluating the enormous amounts of design concepts that can be generated synthetically. In the NLP space, generative models are scored against metrics such as BLUE [55], ROUGE [56], and perplexity [57] to intrinsically evaluate the quality of generated text for tasks such as machine translation or summarization. However, as discussed by Regenwetter *et al.*, these statistical evaluation metrics for generative models do not translate well for engineering design problems [58], and in their paper, the authors provide a review of other possible metrics that could be used to evaluate design concepts. Relevant to the current work are the "similarity" and "distribution matching" metrics, which enable the comparison of a generated dataset to a human dataset in terms of realism and coverage, and the "novelty" and "diversity" metrics, which allow the comparison of various generated datasets in terms of how much design space they cover. Nonetheless, prior work around conceptual generation of design concepts, as discussed in Section 2.1, has leveraged a different set of metrics to evaluate novelty, including auxiliary ML binary classification models to evaluate relevancy of the concepts to the prompt [15], Word Mover's Distance (WMD) [36], as well a distance metric derived from embeddings in a semantic network [36, 59].

Our work differs from prior literature in that we first leverage expert evaluation as a baseline for our task on a subset of our data, and second, we leverage appropriate computational metrics that allow us to scale our results to a larger sample of generated design concepts, as described later in Section 3.4.2.

## 3 METHOD

This section gives an overview of the design prompts used to generate the data (Section 3.1), the methodology used to generate a set of base solutions using GPT-3 (Section 3.2), the prompt engineering performed to generate improved solutions (Section 3.3), and the metrics used to compare the expert and computational evaluations for both generated and crowdsourced design solutions (Section 3.4).

### 3.1 Design Prompts

In this paper, we used GPT-3 to generate solutions based on 12 design problems shown in Table 1. These design problems were taken from a previous paper that analyzed whether crowdsourcing data can be useful for generating inspirational stimuli [8]. We also used crowdsourced data from the same paper written by Goucher-Lambert *et al.* [8] to make baseline comparisons with the GPT-3 generated design solutions.

The problems were modified slightly to allow for GPT-3 to generate 100 design solutions to match the 100 crowd-sourced design solutions received from the previous paper. By only using the design prompts originally used to get crowdsourced design solutions, GPT-3 would only generate one solution because there was no specification as to how many design solutions the GPT-3 model should generate. Therefore, the phrase "Generate 100 design solutions for" was attached to the beginning of each design problem, as shown in Figure 1, to enforce GPT-3 to generate 100 design solutions. Additionally, if a period was added at the end of the design problem, GPT-3 generated solutions that were very brief and nondescript. This is contrary to the very de-scriptive solutions generated from the crowdsourcing platform. However, if a period was removed at the end of the design problem, GPT-3 generated solutions that were more similar in length to the crowdsourced solutions. Therefore, terminal punctuation was removed from every prompt.

**TABLE 1**: The list of design prompts used to generate the design solutions with the LLM.

| ID | Problem |
|---|---|
| 1 | Generate 100 design solutions for a lightweight exercise device that can be used while traveling |
| 2 | Generate 100 design solutions for a lightweight exercise device that can collect energy from human motion |
| 3 | Generate 100 design solutions for a new way to measure the passage of time |
| 4 | Generate 100 design solutions for a device that disperses a light coating of a powdered substance over a surface |
| 5 | Generate 100 design solutions for a device that allows people to get a book that is out of reach |
| 6 | Generate 100 design solutions for an innovative product to froth milk |
| 7 | Generate 100 design solutions for a way to minimize accidents from people walking and texting on a cell phone |
| 8 | Generate 100 design solutions for a device to fold washcloths, hand towels, and small bath towels |
| 9 | Generate 100 design solutions for a way to make drinking fountains accessible for all people |
| 10 | Generate 100 design solutions for a measuring cup for the blind |
| 11 | Generate 100 design solutions for a device to immobilize a human joint |
| 12 | Generate 100 design solutions for a device that can help a home conserve energy |

### 3.2 GPT Base Solutions

Design solutions were generated using the OpenAI playground and command-line interface (CLI). Within the OpenAI playground, the user can choose from several models, and we chose to use *davinci-003* as it is reported to be the higher-performing model at the time of writing.

The output of the generated text can also be controlled by several parameters that impact the generation of the text. These parameters include *temperature*, *top-p*, *frequency penalty*, and *presence penalty*. *Presence penalty* and *frequency penalty* control the generations' repetitiveness, and *temperature* and *top-p* control the randomness of the output. In this paper, we kept all the default parameters as suggested by OpenAI, but we chose to set the *temperature* parameter to 0.9 to diversify the output of the generated solutions, as done in [14]. The generated data and supporting code are publicly available[1].

## 3.3 GPT Prompt Engineering

In order to gain more insight into how prompt engineering affects the LLM-generated design solutions, we employ zero-shot and few-shot learning methods [11, 60], as shown in Figure 1, and we quantitatively evaluate the outputs as described in Section 3.4.2.

**3.3.1 Zero-Shot Method** GPT-3 is capable of performing zero-shot learning [11], a technique used in NLP that allows a model to perform a task that it was not explicitly trained to do [61]. In our case, while the model we use was not trained on a design concept generation task, we refer to the standard prompt fed into the model as the **base prompt** (Figure 1(a)), while we refer to the modified prompt used to guide the model towards different solutions as the **zero-shot prompt** (Figure 1(b)). To create the zero-shot prompts, we take the 12 base prompts and modify them to include descriptive adjectives *novel*, *diverse*, and *unique*, to guide GPT-3 to generate solutions different from the base prompt. For example, as shown in Figure 1, the base design prompt has the phrase "Generate 100 design solutions for a lightweight exercise device that can be used while traveling"; however, the inclusion of the adjective *unique* transforms the base prompt to "Generate 100 *unique* design solutions for a lightweight exercise device that can be used while traveling".

**3.3.2 Few-Shot Method** Few-shot learning methods have been found to improve task-agnostic performance of large language models [60]. Without the need to retrain or fine-tune the model on a domain-specific corpus, few-shot learning is a method by which the prompt is augmented with one or a few examples of the task [61]. For our task, as shown in Figure 1(c), we append to the base prompt an example of a design solution, and then feed the expanded prompt into the model. For each of the 12 design prompts, we sample 5 random solutions from the human dataset to generate 5 different solution sets, to prevent one sampled human prompt from dominating the resulting generated set. We then evaluate the solutions based on our computational metrics for similarity to the whole human dataset and novelty and report the mean and standard deviation over the 5 solutions sets.

## 3.4 Evaluation Metrics

The design output from both GPT-3 generated solutions and crowdsourced solutions were examined in order to determine their differences.

**3.4.1 Expert Evaluation** The metrics used to evaluate the design solutions were derived from literature [8, 42, 62]. The quality measure is subjective by nature, and there were frequent disagreements over the quality of the design solutions between expert evaluators. In addition, when comparing two expert evaluations, prior work [8] has shown that the quality measure had the lowest inter-rater reliability score; therefore, we chose not to do expert evaluation of the quality measure and evaluated the solution output using only measures of feasibility, novelty, and usefulness. The following characteristics of the solutions were explored:

1. **Feasibility**: rated on an anchored scale from 0 (the technology does not exist to create the solution) to 2 (the solution can be implemented in the manner suggested).
2. **Novelty**: rated on an anchored scale from 0 (the concept is copied from a common and/or pre-existing solution) to 2 (the solution is new and unique). Of note, "novelty" is considered to be the uniqueness of the solution with respect to the existing design space and with respect to the entire generated solution set.
3. **Usefulness**: rated on an anchored scale from 0 (completely off-topic and not related to the solution at all) to 2 (the solution is helpful given the context of the prompt).

Two experts, both specializing in design theory and methodology, were trained to perform all ratings for both GPT-3 and crowdsourced design solutions. Consistency was assessed over a subsample of the data using the Cohen's Kappa test, and a subsample of 20 solutions were evaluated for each design prompt and mode of generation (for a total of 120 designs) to assess for a fair-moderate reliability of correlation between the two evaluators. A total of 600 design solutions were evaluated, and a One-Way ANOVA was performed to test for statistically significant differences across feasibility, novelty, and usefulness between GPT-3 generated design solutions and AmazonTurk crowdsourced design solutions.

**3.4.2 Computational Evaluation** To overcome the cost limitations that come with expert human evaluations, we implement two computational metrics from [58] to evaluate the output of the generative models: the mean of the nearest generated samples and the convex hull hypervolume of the solution sets.

Our chosen computational evaluation metrics necessitate the ability to calculate the distance between two designs. As both our crowdsourced dataset and the generated datasets are solely composed of text data, we leverage a pre-trained textual embedding network, SentenceBERT [37], from Sentence-Transformers [29] to embed each individual design solution into a 384-dimensional vector. We use the general purpose model `all-MiniLM-L6-v2` for its balance of speed, size, and quality.

As a measure of similarity and to evaluate the distribution of proposed solutions across the hyper-dimensional design space,

---

[1] https://github.com/kevinma1515/gpt_IDETC

we use the **nearest generated sample** metric. This allows us to capture the coverage of the generated set of solutions compared to the human set of solutions. To calculate it, we measure the cosine distance between each human solution to the nearest generated solution and take the mean of all distances to evaluate the whole set.

To measure the average novelty and diversity of the solution sets as a whole, we use the **convex hull volume** metric. Comparing the total volume of the generated sets to the crowdsourced set allows us to measure the spread of the solutions across the hyper-dimensional design space. This measure was found to be correlated with human evaluation of novelty in low-dimensional design problems [63]. Thus, we first project the 384-dimensional embedding vectors down to a 3-dimensional space using principal component analysis (PCA). We do this to increase the relative volume and density of the embedding spaces, and validate this method by monitoring the variance described by the first three principal components. Then, we calculate the volume of each solution set.

## 4 RESULTS AND DISCUSSION

In this section, we answer our research questions, and evaluate how GPT-3 generated solutions differ from the crowdsourced design solutions. We also explore how prompt engineering can improve the results of GPT-3 generated solutions through computational evaluations of similarity and novelty.

### 4.1 Expert Evaluation

Solutions were rated using the methods outlined in Section 3.4.1. Of the 12 prompts presented in in Table 1, we evaluated only prompts 1, 3, and 4 because early exploratory study revealed that these prompts yielded longer responses, which provides more design details for the experts to evaluate. Both raters evaluated a random subset of solutions from 20 participants for each category of design prompt (120 designs per prompt) across the sub-dimensions of interest (Feasibility, Novelty, and Usefulness) and consistency was assessed using the Cohen's Kappa Score. A moderate level of correlation was obtained for 4 of the 18 evaluations (Cohen's Kappa $> 0.41$) and a fair level of correlation was obtained for the rest (Cohen's Kappa $> 0.21$). Level of agreement between the two raters indicate minimal to weak level of agreement that exposes the limitation in the use of expert raters to evaluate design solutions, which will be explained in further detail in Section 4.3 [64]. For the purpose of this paper, we will proceed to use the human evaluations because it reveals interesting trends about RQ1 that coincide with the computational evaluations done in Section 4.2.

These results evaluate whether there are any statistically significant differences between the means for each measurement between GPT-3 and crowdsourced solutions as reported from Figure 2. For the feasibility, novelty, and usefulness rating, the ratings showed a statistically significant difference between generated design solutions and crowdsourced design solutions for prompts 1, 3, and 4 (p-value $< 0.01$ for all prompts in all 3 metrics). As seen in Figure 2, generated solutions received much higher average feasibility and usefulness ratings than crowdsourced solutions. However, generated solutions received much lower average novelty ratings than crowdsourced solutions. Of note, is the high standard deviation between all bar plots in Figure 2. This happens because there is a high degree of variance in the quality of the solutions based on novelty, feasibility, and usefulness for both GPT-3 generated design solutions and crowdsourced design solutions.

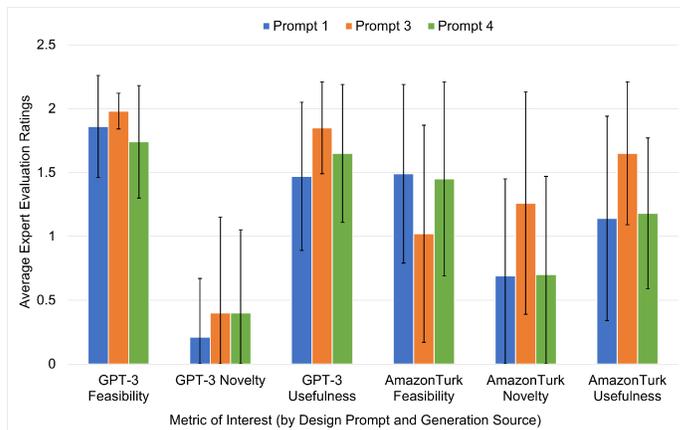

**FIGURE 2**: The average expert evaluation ratings between different sources of generation and design prompts. Each prompt is color coded and the x-axis indicates the category for each metric and the source of the generated design solution (GPT-3 versus crowdsourced from AmazonTurk).

The results laid out in Figure 2 are a product of the evaluation rubric described in Section 3.4.1. Feasibility and usefulness are based on how realistic the solutions are and how closely the solutions address the problem posed in the prompt. Since GPT-3 is trained on text databases scraped from the internet [11], the solutions will closely address the prompt with realistic solutions scraped from the internet unless the model is explicitly told to generate solutions that are very different from existing design solutions. On the other hand, design solutions from crowdsourcing platforms are generated by humans that are not necessarily limited in their imagination and creativity. Humans are capable of generating solutions that would be considered off-topic or infeasible; therefore, their solutions are more likely to be less realistic and less related to the prompt, which explains their lower average scores in feasibility and usefulness as seen in Figure 2.

Likewise, as described from the rubric in Section 3.4.1, the novelty score measures not only the novelty of the solution with respect to the existing design space but also with respect to the entire generated solution set. As mentioned in Section 3.1, we provided inputs into GPT-3 that would generate 100 design solutions for each design problem; however, after a couple of design solutions, GPT-3 would frequently repeat the same solution but with slight modifications in adjectives and features. This is a major limitation to GPT-3 that has been described in [11], where Brown *et al.* noted that GPT-3 will repeat itself semantically until it starts to lose coherence over sufficiently long passages (inputs), causing the model to contradict itself. Design solutions from crowdsourcing platforms do not face this problem because, generally, each crowdsourced worker comes from a different background, so most of the design solutions are quite distinct from one another. However, solutions from crowdsourcing platforms still face the same issue of generating solutions that already exist in the current design space, especially if the worker was not prompted to generate new design solutions. As a result, on average, design solutions generated by crowdsourcing platforms have a higher novelty than the ones from GPT-3 as seen in Figure 2.

**TABLE 2**: GPT-3 generated and crowdsourced (crowd) concept examples according to expert ratings on feasibility (F), novelty (N), usability (U) ratings across the 3 prompts (ID).

| F | N | U | ID | Source | Response |
|---|---|---|----|--------|----------|
| 2 | 0 | 2 | 4 | Crowd | Use the same container that weed killer comes in or bug spray and apply the powder there |
| 0 | 2 | 0 | 3 | Crowd | Non electric dog containment system. Two 50' strands of paracord so the the paracord will meet at a central location about 10' above ground. This will use zip line technology to safely keep dog in yard. An 8' leash will be needed to directly hook on to dogs harness so that it stays in the most vertical line as possible for safety reasons. |
| 0 | 1 | 2 | 1 | Crowd | A pair sturdy (empty) balloons that you fill up from your hotel bathroom water tap that connect together with a strong, telescoping bar. |
| 1 | 2 | 1 | 4 | GPT-3 | 5. A wall-mounted bird-cage-style powder dispenser with a motorized rotor. |
| 2 | 1 | 2 | 3 | GPT-3 | 14. Digital clock with a set of spinning cubes that move with each passing hour |
| 2 | 0 | 2 | 1 | GPT-3 | 14. A handheld weighted bar for strength exercises |

Another feature to note from Figure 2 is the comparatively higher variance for crowdsourced design solutions than GPT-3 generated design solutions. The high degree of variance can be observed in Figure 2 where the error bars for crowdsourced solutions are much larger than the ones for GPT-3 generated solutions. GPT-3 generates design solutions strictly based on the inputs it is provided and what data GPT-3 was trained on [11]. On the contrary, crowdsourced workers come from different backgrounds and prior knowledge, so they can provide solutions that range from very unrealistic to very practical or provide solutions that are either very on-topic or very off-topic. Some examples of the generated concepts can be seen in Table 2, and all generated and crowdsourced concepts can be found on our repository[1]. Overall, these differences in average scores for feasibility, usefulness, and novelty, along with the differences in their respective standard deviations, reveal how there is a large discrepancy in the characteristics of generated solutions from GPT-3 and crowdsourcing platforms.

### 4.2 Computational Evaluation

**Nearest Generated Sample:** To evaluate the similarity of the large quantity of generated design solutions to the crowdsourced set of solutions, we leverage the mean nearest generated sample metric, as described in Section 3.4.2. We evaluate the similarity of the crowdsourced solution set to the GPT-3 Base, GPT-3 zero-shot, and GPT-3 few-shot sets per design prompt and calculate the mean and standard deviation referenced in Table 3. We find that the GPT-3 few-shot results are, on average, the most similar to the crowdsourced set, with a mean cosine similarity of 0.804 (SD=0.06), while the GPT-3 zero-shot prompt modified with the *unique* adjective yields the lowest mean cosine similarity of 0.746 (SD=0.06).

Calculating the nearest generated sample between GPT-3 and crowdsourced solutions allows us to compare the coverage of the GPT-3 generated set of design solutions relative to the crowdsourced set of solutions. Thus, the nearest generated sample metric allows for a quick evaluation between the two generative methods: LLMs and crowdsourced design solutions. Results reveal that, on average across all prompt engineering methods, the nearest generated solutions between crowdsourced and GPT-3 have a high degree of similarity. However, the results in Table 3 reveals that similarity is highly dependent on the prompt ID and its corresponding prompt engineering conditions.

Broadly speaking, there is no consistency around which design prompt input into GPT-3 leads to solutions most similar to crowdsourced solutions. This implies that GPT-3 is a highly stochastic model that can output solutions which vary greatly based on small changes to the design prompt. Most notably, in design prompt 2, by adding the word *diverse* in its design prompt in a similar manner as shown in Figure 1 and as explained in Section 3.3, GPT-3 generated solutions increased in similarity from 0.627 (SD=0.05) to 0.805 (SD=0.07). On the other extreme, by adding the word *diverse* in design prompt 11, GPT-3 generated solutions decreased in similarity from 0.738 (SD=0.05) to 0.675 (SD=0.06).

As described in Section 3.3, prompt engineering was em-ployed in this paper to gain insight into how prompt engineering could affect LLM-generated design solutions. Past research has shown that when LLMs receive additional reasoning steps as inputs into the model, the ability of LLMs to perform complex reasoning should improve [65]. As a result, we employed the few-shot method, as described in Section 3.3.2, in an attempt to alleviate the highly stochastic behavior of GPT-3. Table 3 reveals that, on average, when the few-shot method was utilized, similarity across all design prompts increased. However, some design prompts, such as design prompt ID 1, 2, 3, 4, 5, 8, 9, and 10, did not improve in similarity from one of the prompt engineering methods used in zero-shot to few-shot, which implies that the GPT-3 model is extremely sensitive to the inputs it receives.

**TABLE 3**: Similarity of generated datasets to the human dataset per design prompt, measured using the nearest generated sample metric. A value of one would indicate that the model generated an equal solution set to the human set. Nearest generated sample is a measure of design space coverage, to ensure all key modalities of the design space are reflected in the generated designs.

| Prompt ID | GPT-3 Base | GPT-3 Zero-shot | | | GPT-3 Few-shot |
|---|---|---|---|---|---|
| | | Unique | Novel | Diverse | |
| 1 | 0.737 (SD=0.05) | 0.728 (SD=0.04) | 0.788 (SD=0.05) | 0.773 (SD=0.06) | 0.773 (SD=0.05) |
| 2 | 0.627 (SD=0.05) | 0.754 (SD=0.04) | 0.694 (SD=0.05) | 0.805 (SD=0.07) | 0.743 (SD=0.07) |
| 3 | 0.727 (SD=0.08) | 0.759 (SD=0.07) | 0.758 (SD=0.09) | 0.747 (SD=0.07) | 0.755 (SD=0.08) |
| 4 | 0.778 (SD=0.10) | 0.716 (SD=0.07) | 0.786 (SD=0.11) | 0.738 (SD=0.08) | 0.746 (SD=0.08) |
| 5 | 0.796 (SD=0.05) | 0.708 (SD=0.05) | 0.796 (SD=0.04) | 0.842 (SD=0.03) | 0.800 (SD=0.03) |
| 6 | 0.807 (SD=0.07) | 0.805 (SD=0.04) | 0.686 (SD=0.07) | 0.784 (SD=0.06) | 0.814 (SD=0.05) |
| 7 | 0.768 (SD=0.01) | 0.782 (SD=0.04) | 0.716 (SD=0.01) | 0.697 (SD=0.06) | 0.789 (SD=0.05) |
| 8 | 0.704 (SD=0.03) | 0.717 (SD=0.03) | 0.741 (SD=0.04) | 0.869 (SD=0.05) | 0.842 (SD=0.03) |
| 9 | 0.770 (SD=0.09) | 0.781 (SD=0.10) | 0.754 (SD=0.10) | 0.816 (SD=0.11) | 0.757 (SD=0.11) |
| 10 | 0.765 (SD=0.04) | 0.833 (SD=0.01) | 0.840 (SD=0.02) | 0.821 (SD=0.02) | 0.820 (SD=0.05) |
| 11 | 0.738 (SD=0.05) | 0.668 (SD=0.07) | 0.708 (SD=0.05) | 0.675 (SD=0.06) | 0.798 (SD=0.04) |
| 12 | 0.753 (SD=0.04) | 0.674 (SD=0.03) | 0.752 (SD=0.05) | 0.776 (SD=0.05) | 0.842 (SD=0.04) |
| Mean | 0.748 (SD=0.06) | 0.746 (SD=0.06) | 0.775 (SD=0.07) | 0.779 (SD=0.07) | 0.804 (SD=0.06) |

**TABLE 4**: Novelty evaluation of each design space measured by the convex hull volume. Measuring the breadth of the design space coverage of each model gives an indication of the diversity in responses generated for each prompt. A higher novelty measure indicated that the solutions for each prompt do not repeat and cover a wide design space.

| Prompt ID | Human | GPT-3 Base | GPT-3 Zero-shot | | | GPT-3 Few-shot |
|---|---|---|---|---|---|---|
| | | | Unique | Novel | Diverse | |
| 1 | 0.249 | 0.324 | 0.332 | 0.255 | 0.348 | 0.214 (SD=0.05) |
| 2 | 0.244 | 0.197 | 0.326 | 0.150 | 0.175 | 0.176 (SD=0.05) |
| 3 | 0.322 | 0.238 | 0.275 | 0.161 | 0.201 | 0.230 (SD=0.08) |
| 4 | 0.317 | 0.193 | 0.347 | 0.149 | 0.190 | 0.186 (SD=0.07) |
| 5 | 0.338 | 0.368 | 0.342 | 0.264 | 0.089 | 0.211 (SD=0.12) |
| 6 | 0.261 | 0.164 | 0.198 | 0.074 | 0.187 | 0.168 (SD=0.07) |
| 7 | 0.330 | 0.308 | 0.207 | 0.245 | 0.348 | 0.273 (SD=0.04) |
| 8 | 0.277 | 0.121 | 0.094 | 0.148 | 0.161 | 0.163 (SD=0.1) |
| 9 | 0.327 | 0.196 | 0.151 | 0.302 | 0.155 | 0.268 (SD=0.1) |
| 10 | 0.307 | 0.400 | 0.175 | 0.105 | 0.191 | 0.167 (SD=0.1) |
| 11 | 0.303 | 0.280 | 0.293 | 0.191 | 0.358 | 0.149 (SD=0.09) |
| 12 | 0.254 | 0.321 | 0.456 | 0.183 | 0.196 | 0.226 (SD=0.08) |
| Mean | 0.294 (SD=0.03) | 0.259 (SD=0.09) | 0.266 (SD=0.1) | 0.186 (SD=0.07) | 0.217 (SD=0.09) | 0.2 (SD=0.04) |

**Convex Hull Volume:** To evaluate the novelty of the generated and crowdsourced solutions sets, we leverage the measure of the convex hull volume as described in Section 3.4.2. The convex hull volumes of all solution sets for the design prompts and models are reported in Table 4. We averaged the volumes over the prompts to generalize the novelty of the different methods. This shows that the crowdsourced solution set has the highest volume with 0.294 (SD=0.03) while augmenting the prompt with the *novel* adjective actually reduces novelty and yields the smallest convex hull volume of 0.186 (SD=0.07).

Prior research has shown that inspirational stimuli are dependent on the analogical distance of the solutions to the design prompt [2, 3, 17]; therefore, generating or crowdsourcing solutions which span across a large hypervolume of the design solution space could be desirable. The convex hull volume metric is intended to measure the breadth of the design space coverage of the solutions set, so it is sensitive to repetitions of solutions in the set (which reduce the volume) as well as unique or distant solutions (which increase the volume). This could explain why, of the generated solution sets, the GPT-3 zero-shot prompt augmented with the *unique* adjective yields the highest novelty as defined in Section 3.4.1, with a convex hull volume of 0.266 (SD=0.1). However, GPT-3 few-shot model and GPT-3 zero-shot model with the *novel* adjective yield the lowest novelty as defined in Section 3.4.1, with a convex hull volume of 0.2 (SD=0.04) and 0.186 (SD=0.07), respectively. A qualitative evaluation of the solution set reveals that these generated solution sets contained repetitions of the same design solutions but with slight modifications in their description, which could explain the low novelty score in both cases. These negative effects were also qualitatively observed in design solutions generated by GPT-3 base design prompts, and the difference in the computational evaluation of novelty, as observed in Table 4, are also similarly reflected in the human evaluation of novelty, as observed in Figure 2.

### 4.3 Limitations and Future Work

While showing promising results, the methods explored in the current work to generate a synthetic set of inspirational design solutions to a prompt still fall behind established methods, such as crowdsourcing. Future work could expand on these methods by fine-tuning LLMs on domain-specific literature, leveraging different models for generating text, and expand on the prompt engineering that began with our zero-shot and few-shot experiments.

Both the human and the computational evaluation metrics could be further refined. While the human evaluation of individual design solutions from experts proved insightful, there remains a question of the value of the solution sets in providing inspirational stimuli to designers. This could be evaluated more directly with a human study, where the design solutions created by subjects prompted by either human or LLM-generated solutions sets are evaluated. Additionally, if human evaluations are to be pursued further, future work should establish a more refined rubric that places the limitations of GPT-3 generated design solutions (e.g., non-descriptive solutions) into consideration. For the computational metrics, future work should verify the effectiveness and validity of the chosen metrics and others in regard to textual design solutions. Additionally, future work should look into exploring new methods of computational measurements. Currently, there exist methods to evaluate diversity, uniqueness, and similarity, but there are no computational methods to evaluate usefulness and feasibility, which currently can only be done through expensive expert evaluations as described in Section 3.4.1. Finally, LLMs are still considered a "black box", and, as mentioned in Section 4.2, their outputs are extremely sensitive to the user's input. Future work should address this issue by investigating how designers can leverage existing prompt engineering techniques [66,67,68] to support early-stage design ideation.

A qualitative evaluation of all design solutions used in this work shows that the generated solutions are shorter and contain less information than the crowdsourced solutions. While this effect is mitigated by our few-shot method, this difference is at the core of why generated solutions did not perform as well as crowd-sourced solutions. Additionally, this paper only examines one LLM, GPT-3, so there still remains a question of whether or not this issue is unique to GPT-3 or ubiquitous amongst all LLMs. Therefore, future work could explore methods for yielding better design responses via prompt engineering [65], and explore other LLMs using the methods described in this paper.

### 5 CONCLUSION

Our long-term motivation is to support designers during the conceptual design stage by using pre-trained LLMs to generate creative design solutions more quickly and cheaply than current methods, such as crowdsourcing. This paper assesses whether design solutions generated by LLMs are fundamentally different from crowdsourced design solutions and uses expert and computational evaluations to evaluate the differences. Results reveal that, with the use of few-shot learning, LLMs are capable of generating design solutions that are similar to crowdsourced solutions, but these modifications lead to a decrease in the diversity of solutions that LLMs are capable of generating. Expert evaluations reveal that LLMs generate solutions that are more *feasible* and *useful* than crowdsourced solutions but less *novel*. This paper provides a foundation for future work to explore the use of LLMs in providing conceptual design ideas to designers. We see many opportunities to expand this work to support the creative aspects of the early-stage conceptual design by leveraging large, pre-trained, and multi-modal ML models, fine-tuned on engineering-specific domain knowledge.

### ACKNOWLEDGMENT

The authors would like to thank Yakira Mirabito for her contribution in assisting with the expert evaluation, and the anonymous Amazon Mechanical Turk workers who provided design solutions.